\pdfoutput=1
\documentclass[11pt,twocolumn,a4paper]{article}
\usepackage{times}
\usepackage{latexsym}
\usepackage{xcolor}  
\usepackage{graphicx}  
\usepackage{url}
\usepackage{authblk}
\usepackage{microtype}

\title{Exploring the Capacity of a Large-scale Masked Language Model to Recognize Grammatical Errors}

\author[1,2]{Ryo Nagata}
\author[3]{Manabu Kimura}
\author[4,5]{Kazuaki Hanawa}
\affil[1]{Konan University}
\affil[2]{JST, PRESTO}
\affil[3]{GRAS Group, Inc.}
\affil[4]{RIKEN AIP}
\affil[5]{Tohoku University}

\date{}

\begin{document}
\maketitle

\section{Introduction}\label{sec:intro}
  Recent studies have shown that masked language models pre-trained on a large corpus (hereafter, simply language models) achieve tremendous improvements over a wide variety of natural language processing (NLP) tasks. These results imply that they are also effective in recognizing erroneous words and phrases known as the task of grammatical error detection. There has been, however, much less work on this aspect of grammatical error detection than in other tasks. One can argue that since language models are trained on language data produced by native speakers of a language (specifically, English in this paper), they might not work well on the present task. This is partly because the target language data are produced by non-native speakers of that language. In other words, English language models do not know at all about grammatical errors made by non-native speakers. Even apart from grammatical errors, the target language is different from the canonical English, meaning that it contains unnatural words/phrases and characteristic language usages that native speakers do not normally use as \cite{nagata13} demonstrate. If so, the effectiveness of language models is not so evident in grammatical error detection.

  Actually, researchers have reported on performance of language models on grammatical error detection and correction, which partly answers the above research questions. \cite{cheng-duan2020} and \cite{kaneko_komachi2019} have shown that BERT-based methods improve grammatical error detection performance in Chinese and English, respectively. \cite{kaneko2020} and \cite{didenko_shaptala2019} have shown a similar tendency in grammatical error correction. While these studies empirically prove the effectiveness of language models in grammatical error detection and correction, the questions why and where language models benefit error detection/correction methods are left unanswered.

  In this paper, we explore this aspect of language models in grammatical error detection to better answer the research questions. We first show that 5 to 10\% of training data are enough for a BERT-based error detection method to achieve performance equivalent to a non-language model-based method can achieve with the full training data. More precisely, recall improves much faster with respect to training data size in the BERT-based method than in the non-language model method while precision behaves similarly. These experimental results suggest that (i) the BERT-based method should have a good knowledge of grammar required to recognize certain types of error and that (ii) it can transform the knowledge into error detection rules by fine-tuning with a few training samples, all of which leads to its high generalization ability in grammatical error detection. Following this, we further show with pseudo error data that it actually exhibits such nice properties in learning rules for detecting errors. For instance, we show that the BERT-based method trained on few (as few as two) instances of a transitive verb with a preposition (e.g., \textit{*discuss about}) can detect the same type of error in other verbs (e.g., \textit{*approach to} and \textit{attend in}). Finally, based on these findings, we explore a cost-effective method for detecting grammatical errors with feedback comments explaining relevant grammatical rules to learners.

\section{Related Work}\label{sec:related_work}
  \cite{rei2017} shows it is useful for neural error detection models to introduce a secondary language model objective together with the main error detection objective. \cite{rei_yannakoudakis2017} compare several other auxiliary training objectives including Part-Of-Speech (POS) tagging and error type identification and find that the language model objective is the most effective. This line of work suggests that grammatical error detection benefits from language modeling although these studies use BiLSTM-based language models instead of masked language models trained on a large corpus.

  As mentioned in Sect.\,\ref{sec:intro}, several researchers have applied masked language models including BERT to grammatical error detection and correction. \cite{cheng-duan2020} and \cite{kaneko_komachi2019} show that error detection methods gain in recall and precision with the use of language models. \cite{bell2019} use BERT-based contextual embeddings for grammatical error detection and compares it with other types of contextual embedding. They show the BERT-based contextual embeddings are effective in almost all error types provided by ERRANT~\cite{bryant} although BERT is not fine-tuned in their study. \cite{kaneko2020} and \cite{didenko_shaptala2019} also show performance improvements in grammatical error correction. To strengthen the findings of these previous studies, we will reveal (at least, partly) why and where error detection methods benefit from language models in the following sections.

  There has been a long history of studies that investigate the linguistic knowledge of language models including the work by~\cite{li2021,ettinger2020,warstadt2020} to name a few. A popular approach is to test whether a language model assigns higher likelihood to the appropriate word than an inappropriate one, given context. The linguistic knowledge to be explored ranges from syntactic/semantic knowledge to common sense. These studies mostly use (i) synthetic test data: sentences that are generated synthetically by using a certain kind of template or (ii) perturbed test data: sentences are generated by perturbing a natural corpus. Our work is different from these previous studies in two points: (i) to our best knowledge, we examine linguistic phenomena that have never been explored before in the conventional studies (e.g., subjects marked with a preposition and errors involving the usages of transitive and intransitive verbs); (ii) we use a real learner corpus with real errors as our test data.%

  \cite{mita2021} examine if an encoder-decoder neural network for grammatical error correction (not BERT-based) can learn the knowledge of grammar from training data for grammatical error correction (pairs of original and corrected sentences). They target five error types: subject-verb agreement, verb form, word order, adjective/adverb comparison, noun number. They use both synthetic and real learner data. They report a negative answer to the research question except for word order errors while their model learns the knowledge to detect the target errors. Our study supports and deepens their findings for a wider variety of error types that are much more difficult to detect (in that it requires a much wider range of linguistic knowledge including POS, lexical, and syntactic knowledge).

\section{Data and Methods}\label{sec:data_and_methods}
\subsection{Real and Pseudo Data}\label{subsec:data}
  In this paper, we use two kinds of data: real and pseudo data. Real data consist of an English learner corpus manually annotated with grammatical errors while pseudo data are automatically generated from a native corpus by perturbing a native English corpus.

  For the real data, we use the data created in the work~\cite{nagata18}. Its base corpus is the written essays in ICNALE~\cite{ishikawa}. It consists of essays written by English learners. Their topics are controlled; they are written on either (a) \textit{It is important for college students to have a part-time job.} or (b) \textit{Smoking should be completely banned at all the restaurants in the country.}, which hereafter will be referred to as \textit{PART-TIME JOB} and \textit{SMOKING}, respectively. Each essay is manually annotated with errors and feedback comments explaining their relevant grammatical rules in detail. These two sources of information help us investigate error types that the BERT-based method can recognize. The original data provide feedback comments concerning preposition errors and more general writing items. In this paper, we limit ourselves to the essays annotated with preposition feedback comments so that we can conduct a deep analysis targeting a class of grammatical errors. Having said that, the target preposition errors involve a much wider range of errors than in the conventional definition of preposition errors (such as the one provided by ERRANT). For instance, the preposition errors in the work~\cite{nagata18} include deverbal prepositions (e.g., \textit{*include} $\to$ \textit{including}), intransitive verbs with a direct object (e.g., \textit{*agree it} $\to$ \textit{agree with it}), a verb phrase used as a noun phrase (\textit{*Lean English is difficult.} $\to$ \textit{To learn/Learning English is difficult.}), and comparison between a phrase and a clause (e.g., \textit{*because an error} $\to$ \textit{because of an error}); see their work for the details.

  The essays are randomly split into training, development, and test sets in the ratios of 85\%, 7.5\%, and 7.5\%, respectively. Table~\ref{tab:stats} shows their statistics\footnote{The data development in the work of \cite{nagata17}. For this work, we obtained data that had not been open to the public yet from the developer.}. To investigate the relationship between the number of training sentences and detection performance, we randomly sample 100, 300, 500, 1000, 3000, 5000, 10000, and all sentences (12,163 and 12,312 in PART-TIME JOB and SMOKING, respectively), resulting in eight sets of training data for each topic. Note that these training, development, and test sets contain error-free sentences.%

\begin{table*}[t]%
\begin{center}
\begin{tabular}{l|ccc|ccc}
\hline
Topic & \multicolumn{3}{c}{PART-TIME JOB} & \multicolumn{3}{|c}{SMOKING} \\
Split & Training &  Dev.\ &  Test  & Training &  Dev.\ &  Test \\
\hline
\# sentences & 12,163  & 1,129  & 1,042  & 12,312 &  1,160 & 1,023 \\
\# tokens       & 205,355 & 18,276 & 17,192	& 201,304 &  18,242 & 17,318 \\
\# feedback comments & 2,439  &	244 & 222     & 2,342  & 230   & 212  \\
\hline
\end{tabular}
\end{center}
\caption{Statistics on Real Dataset.}\label{tab:stats}
\end{table*}%

  For the pseudo data, we use the 1998-2000 New York Times in the AQUAINT Corpus of English News Text~\cite{aquaint} as a base corpus. We automatically generate erroneous sentences by injecting errors into them (one error per sentence). We first obtain chunks and parses by using Spacy\footnote{\url{https://spacy.io/}}. Here, we only use sentences whose lengths are longer than three tokens and shorter than 26 so that we can get reliable chunks and parses. We then add, remove, or replace a word in the sentences based on the analyses.

  While we target all errors labeled as preposition errors in the real data, we only target the following five error types in the pseudo data:%
\begin{description}
\setlength{\parskip}{-0.7mm}
\item[Prepositional infinitive:] \textit{to}-infinitive with other prepositions than \textit{to}.\\
(e.g., \textit{a book to read} $\to$ \textit{*a book for read})
\item[Subject verb:] Verb phrases used as a subject\\
(e.g., \textit{*Lean English is difficult.})
\item[Prepetition + subject:] Subjects used with a preposition\\
(e.g., \textit{*In the restaurant serves good food.})
\item[Transitive verb + preposition:] Transitive verbs used with a preposition\\
(e.g., \textit{*We discussed about it.})
\item[Intransitive verb + object:] Intransitive verbs taking a direct object\\
(e.g., \textit{*We agree it.})
\end{description}
These five error types are selected with the two criteria: (i) they are major errors in the real data; (ii) we can easily write a software program to generate pseudo errors based on chunks and parses. For example, we can find a subject of a sentence from its parse and then can add a randomly-chosen preposition before the subject noun phrase in \textit{*In the restaurant serves good food}. We randomly choose one of the following five prepositions: \textit{at}, \textit{about}, \textit{to}, \textit{in}, and \textit{with} for addition and replacement; an exception is that we only use \textit{for} for Prepositional infinitive (e.g., \textit{a book to read} $\to$ \textit{*a book for read}), which often appears in the real data. Similarly, we can extract pairs of a verb and its direct object from parses and then can add put one of the prepositions before the direct object noun phrase as in \textit{discuss the matter} $\to$ \textit{*discuss about the matter}. Tables~\ref{tab:transitive_verbs} and~\ref{tab:intransitive_verbs} show the target transitive and intransitive verbs, respectively. It should be emphasized that as shown in the tables, there is no overlap of verbs in the training and test data. This means that error detection methods must learn to detect these two types of error in a set of verbs from those in another set of verbs.%

\begin{table}[t]%
\begin{center}
\begin{tabular}{c|c}
\hline
Training/Dev\. Data & Test Data \\
\hline
answer & approach \\
attend & consider \\
discuss & enter \\
inhabit & marry \\
mention & obey \\
oppose & reach \\
resemble & visit \\
\hline
\end{tabular}
\end{center}
\caption{Target Transitive Verbs.}\label{tab:transitive_verbs}
\end{table}%

\begin{table}[t]%
\begin{center}
\begin{tabular}{c|c}
\hline
Training/Dev.\ Data & Test Data \\
\hline
agree &  apply \\
belong & graduate \\
disagree &       listen \\
relate & specialize \\
 &       worry \\
\hline
\end{tabular}
\end{center}
\caption{Target Intransitive Verbs.}\label{tab:intransitive_verbs}
\end{table}%

  From the resulting pseudo error data, we randomly sample $2^k (1 \leq k \leq 10)$ sentences for each error type, resulting in ten sets of training data (e.g., when $k=1$, the set comprises two instances of each error type, ten instances in total. We use these training sets to estimate the relationship between the number of training instances and detection performance. For a validation set, we randomly sample 200 sentences for each error type. Similarly, we use a test set consisting of 200 sentences randomly sampled for each error type plus another 200 error-free sentences. The validation and test sets are fixed regardless of the training data.

\subsection{Grammatical Error Detection Methods}\label{subsec:methods}
  This subsection describes the three methods to be explored and compared. Before looking into them, let us define grammatical error correction formally. Grammar error detection can be solved as a token classification problem\footnote{More generally, it can also be solved as a sequence labeling problem using for example CRF. However, \cite{rei2017} shows that the grammatical error detection task does not benefit from CRF. We actually observed the same tendency in our datasets. Accordingly, we solve it as a token classification problem (without CRF).}. To formalize it, we will denote a sequence of words\footnote{Here, we use the term \textit{words} abstractly to mean word-like objects, which can be words or subwords.} and its length by $w_{1}, \dots, w_{i} \dots, w_{N}$ and $N$, respectively. We will denote the corresponding sequence of labels by $l_{1}, \ldots, l_{i}, \ldots, l_{N}$ where $l_i$ corresponds to the label of $w_{i}$. We assume two sets of labels: (i) either C or E denoting \textit{correct} or \textit{erroneous} in the real data, respectively; (ii) $K$ labels for $K$ error types plus C for \textit{correct} in the pseudo data. Then, grammatical error detection is defined as a problem of predicting the optimal label sequence given $w_{1}, \dots, w_{i} \dots, w_{N}$.

  Basically, we use neural networks to predict the optimal label sequence. In this paper, training is repeated five times with different (but fixed) random seeds. The reported performance values (i.e., recall, precision, and $F_{1.0}$) are averaged over the five runs. Training epochs are ten at the maximum and we adopt the epoch achieving the best $F_{1.0}$ on the development set. Table~\ref{tab:hyper_params} shows the other major hyper parameters\footnote{When we use the pseudo data for training, the number of training sentences can be as small as ten, and we use a rather small batch of five; otherwise we use 32.}.%

\begin{table}[t]%
\begin{center}
\begin{tabular}{c|c}
\hline
Batch size & 5 or 32  \\
Optimization & Adam with decoupled weight\\
 &decay regularization \\
Learning rate & $5e$-$5$, (0.9, 0.999) \\
Epsilon &$1e$-$8$ \\
Weight decay &  $1e$-$2$ \\
\hline
\end{tabular}
\end{center}
\caption{Major Hyperparameters.}\label{tab:hyper_params}
\end{table}%

\subsubsection{BERT-based Method}\label{subsubsec:bert-based_method}
  The BERT-based method takes as input a word sequence $w_{1}, \dots, w_{i} \dots, w_{N}$ and conducts the following procedures:%

\begin{description}
   \item[\textbf{(1) Subword:}] put all $w_{i}$ into their corresponding subwords: $s_{1}, \dots, s_{j} \dots, s_{M}$. Note that the total number of all subwords are generally different from that of all words in the input word sequence.

   \item[\textbf{(2) Encode:}] encode all $s_{j}$ into BERT embeddings $b_{j}$ by:
   \begin{equation}
   b_{j} = \mbox{BERT}(s_{j})
   \end{equation}
   where $\mbox{BERT}(\cdot)$ denotes BERT taking subwords as input and outputs their corresponding embedding vectors of $h$-dimension (specifically, $h=768$ for the BERT base model) from the final layer. We use the BERT base model (uncased) for $\mbox{BERT}(\cdot)$. 

   \item[\textbf{(3) Token classification:}] output the optimal labels by:
   \begin{equation}
    l_{i} = \arg\max\mbox{softmax} (Wb_{j})
   \end{equation}
   where $W$ is a $k\times h$ weight matrix where $k$ is either $2$ or $K$ (the number of different labels). To take care of the difference in the lengths of the input word sequence and the corresponding subword sequence, only the first subword of each word is considered in training and prediction.
\end{description}

\subsubsection{Methods to Be Compared}\label{subsubsec:method2compre}
  For comparison, we select a BiLSTM-based error detection method. Basically it follows the above steps (1) to (3). The major difference is that we use BiLSTM as an encoder in place of BERT. Also, the input word sequence is turned into a sequence of embedding vectors where each embedding vector consists of the concatenation of the conventional word embedding and a character-based embedding. The character-based embedding is obtained by another BiLSTM taking the characters of each word following the work~\cite{akbik2018}. The concatenated embeddings are put into the encoder BiLSTM to produce vectors for prediction in the step (3). Specifically we use the implementation FLAIR~\cite{akbik2019}. We will refer to this method for comparison as the BiLSTM-based method, hereafter.

  We also investigate how effective the fine-tuning of BERT is. Namely, the BERT part of the BERT-based method is fixed during training and the only output layer is adjusted by the training data. We will refer to this method as the BERT-based method without BERT train, hereafter.

\section{Performance on Real Data}\label{sec:effects_on_where2comment}
\begin{figure*}[t]
\centering
\begin{center}
\begin{picture}(130, 177)
    \put(-187, -7){\includegraphics[scale=0.57]{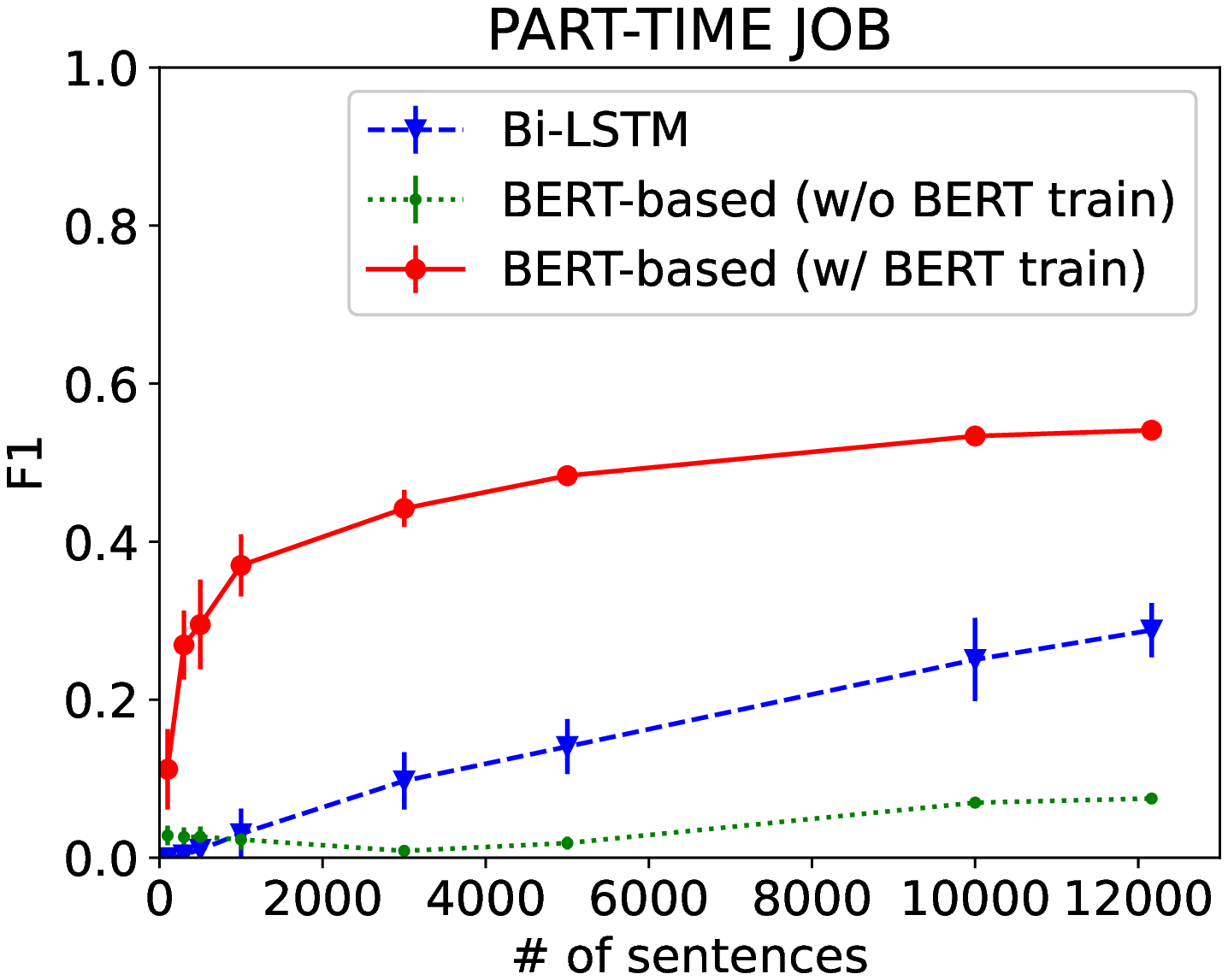}}
    \put(57, -7){\includegraphics[scale=0.57]{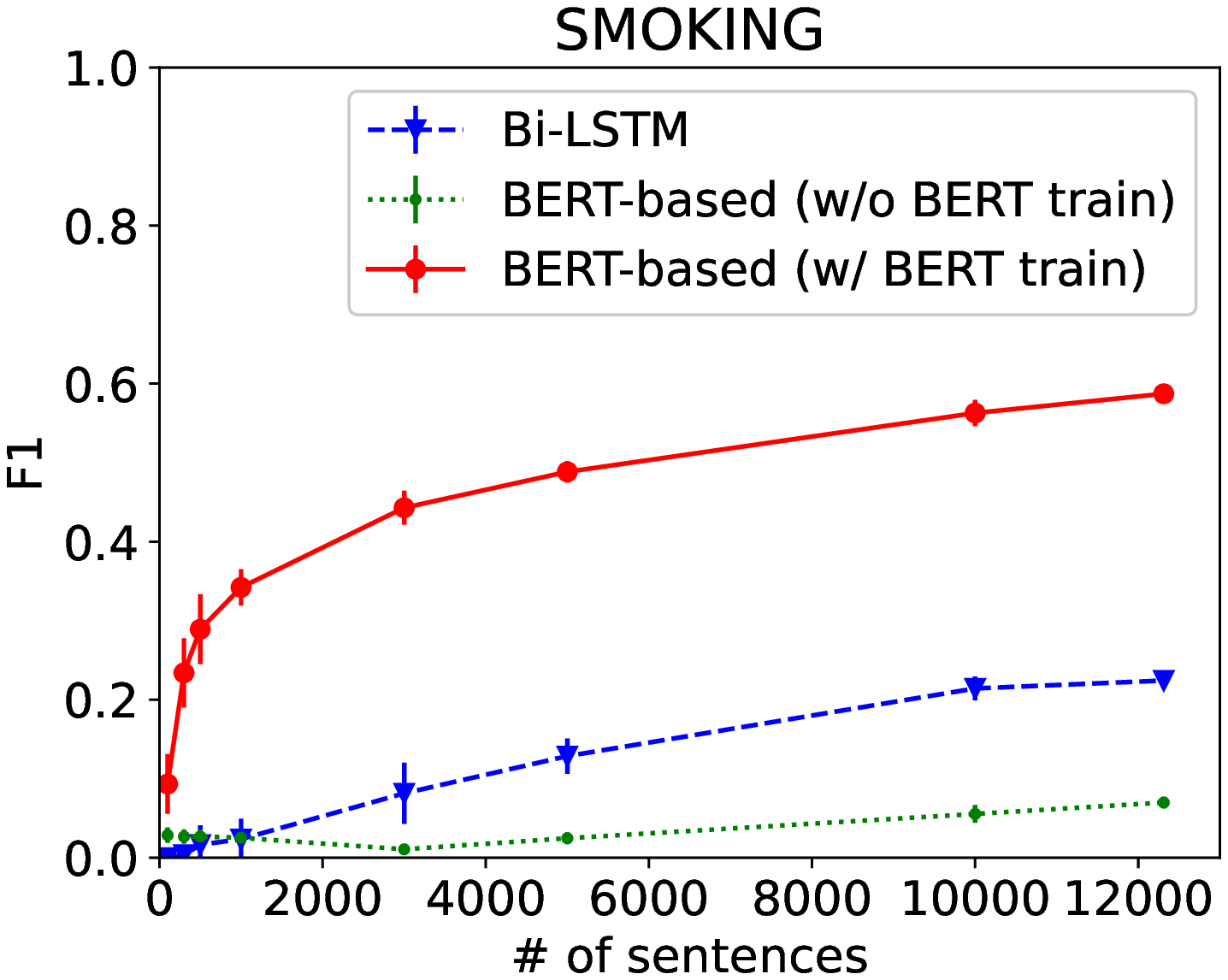}}
\end{picture}
\caption{Detection Performance ($F_{1.0}$) in Relation with Training Size: in-domain test data (average over five runs).}\label{fig:f1_in_domain}
\end{center}
\end{figure*}

  Figure~\ref{fig:f1_in_domain} shows the relationship between the size of training data and $F_{1.0}$ where the size is measured by the number of sentences. The three methods are trained with the specified amount of the data in either topic (PART-TIME JOB or SMOKING) and are tested on the data in the same topic (in-domain test).

\begin{figure*}[t]
\centering
\begin{center}
\begin{picture}(130, 177)
    \put(-187, -7){\includegraphics[scale=0.57]{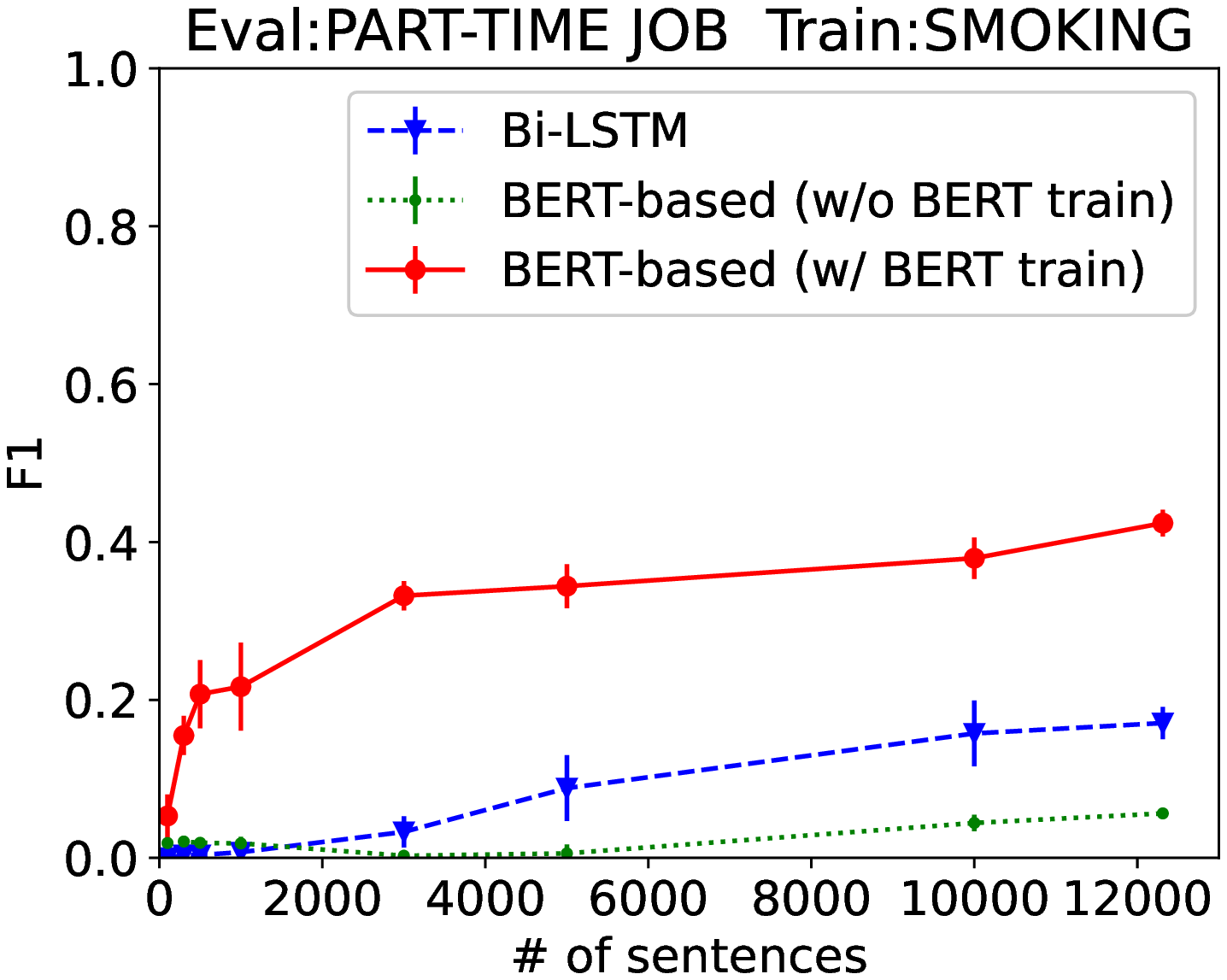}}
    \put(57, -7){\includegraphics[scale=0.57]{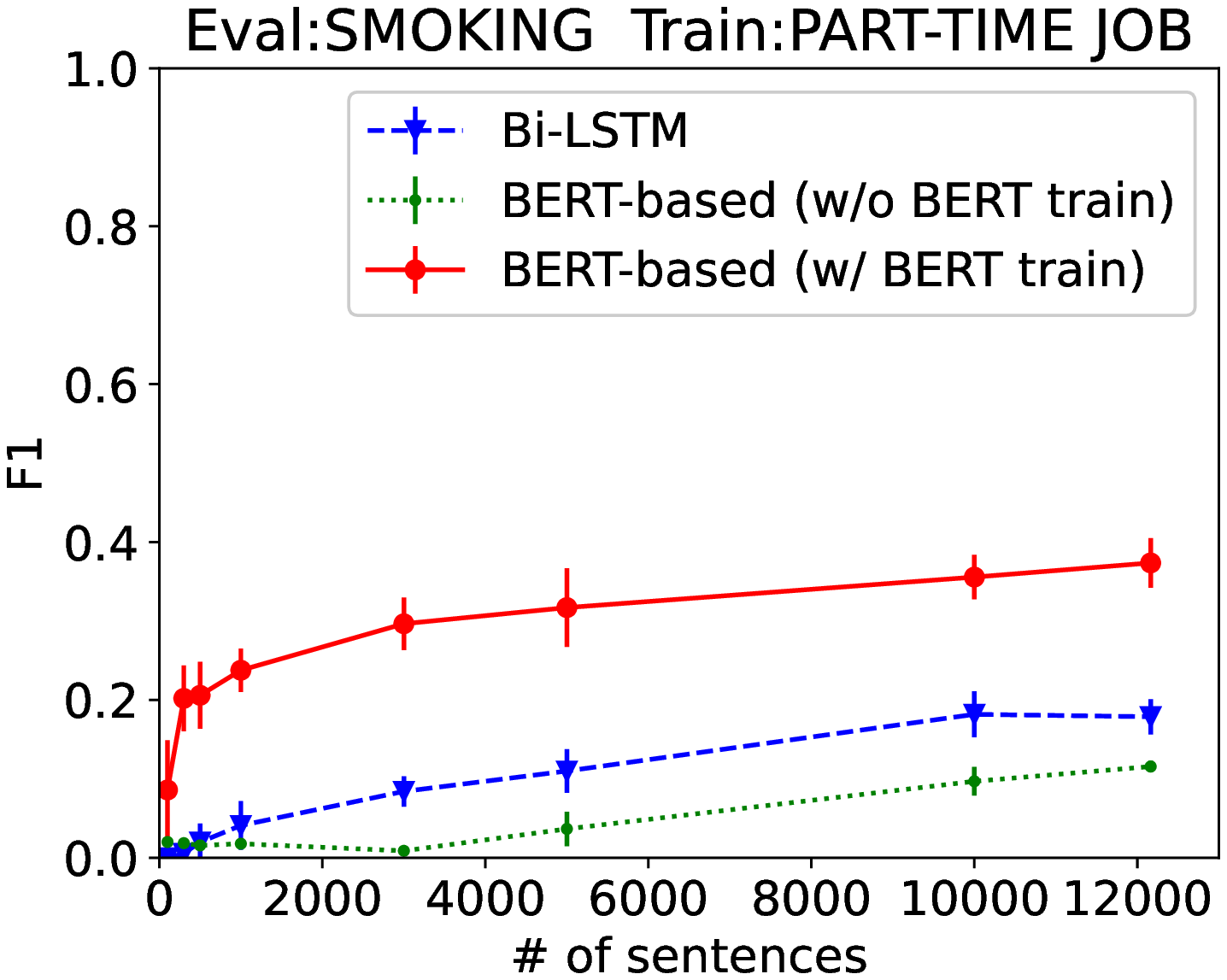}}
\end{picture}
\caption{Detection Performance ($F_{1.0}$) in Relation with Training Size: out-domain test data (average over five runs).}\label{fig:f1_out_domain}
\end{center}
\end{figure*}

  Figure~\ref{fig:f1_in_domain} reveals that the BERT-based method exhibits a performance saturation at a point of 1,000-2,000 training sentences while the BiLSTM-based method almost linearly improves as the number of training sentences increases within the range of the available training data. The graph for the BERT-based method without BERT train exhibits a similar shape to that of the BiLSTM-based method, but the values of $F_{1.0}$ are much lower. In addition, Figure~\ref{fig:f1_in_domain} suggests that it achieves with only 500 to 1,000 training sentences an $F_{1.0}$ equivalent to the $F_{1.0}$ the BiLSTM-based method can achieve with the full training data. Figure~\ref{fig:f1_out_domain} shows the same tendencies in the out-domain test setting. 

  These results support the hypothesis that the masked language models trained on a large corpus have a much higher generalization ability in grammatical error detection. In addition, they empirically show that fine-tuning is crucial in the application of BERT to the task of grammatical error detection.

\begin{figure*}[t]
\centering
\begin{center}
\begin{picture}(130, 177)
    \put(-177, 0){\includegraphics[scale=0.55]{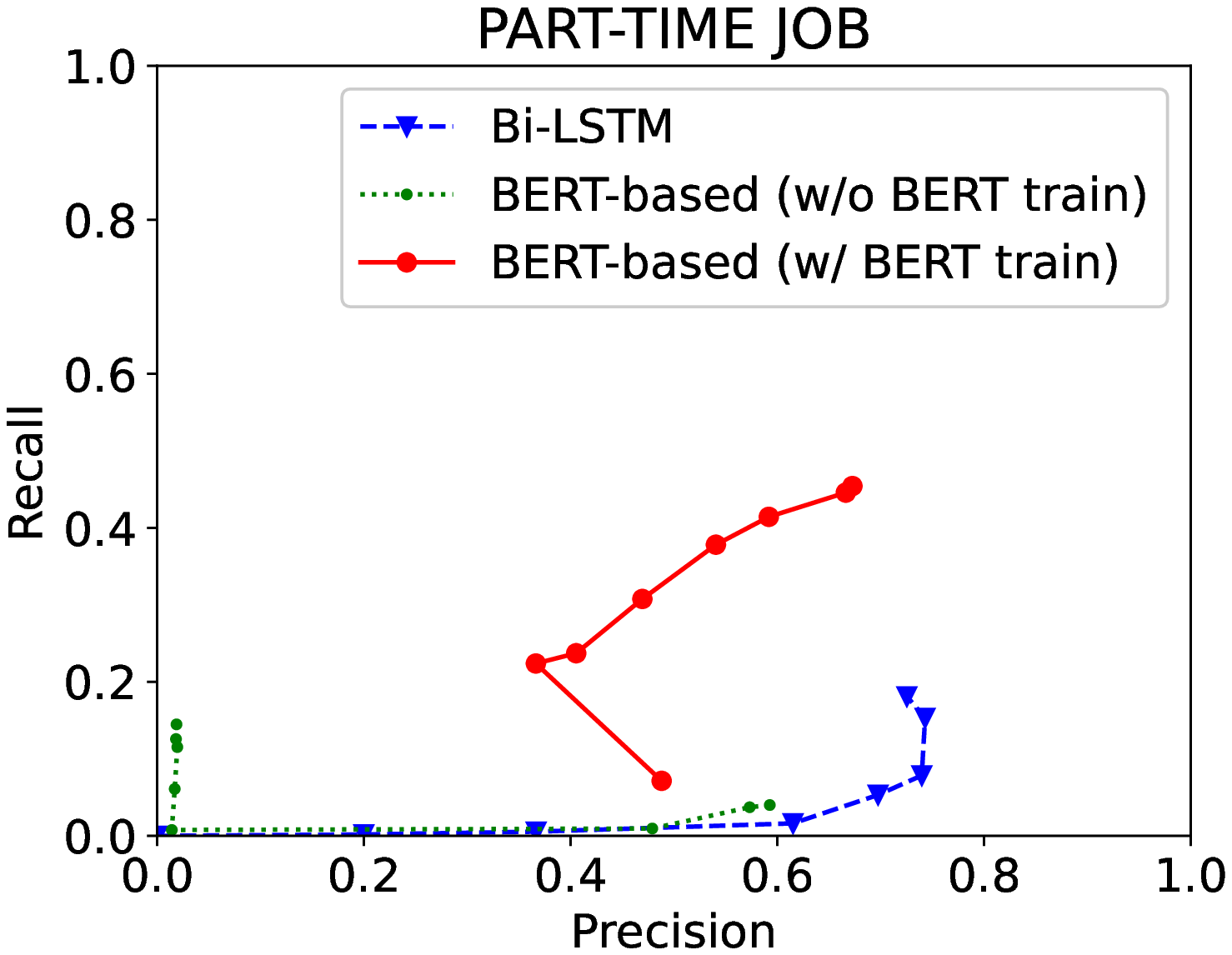}}
    \put(70, 0){\includegraphics[scale=0.55]{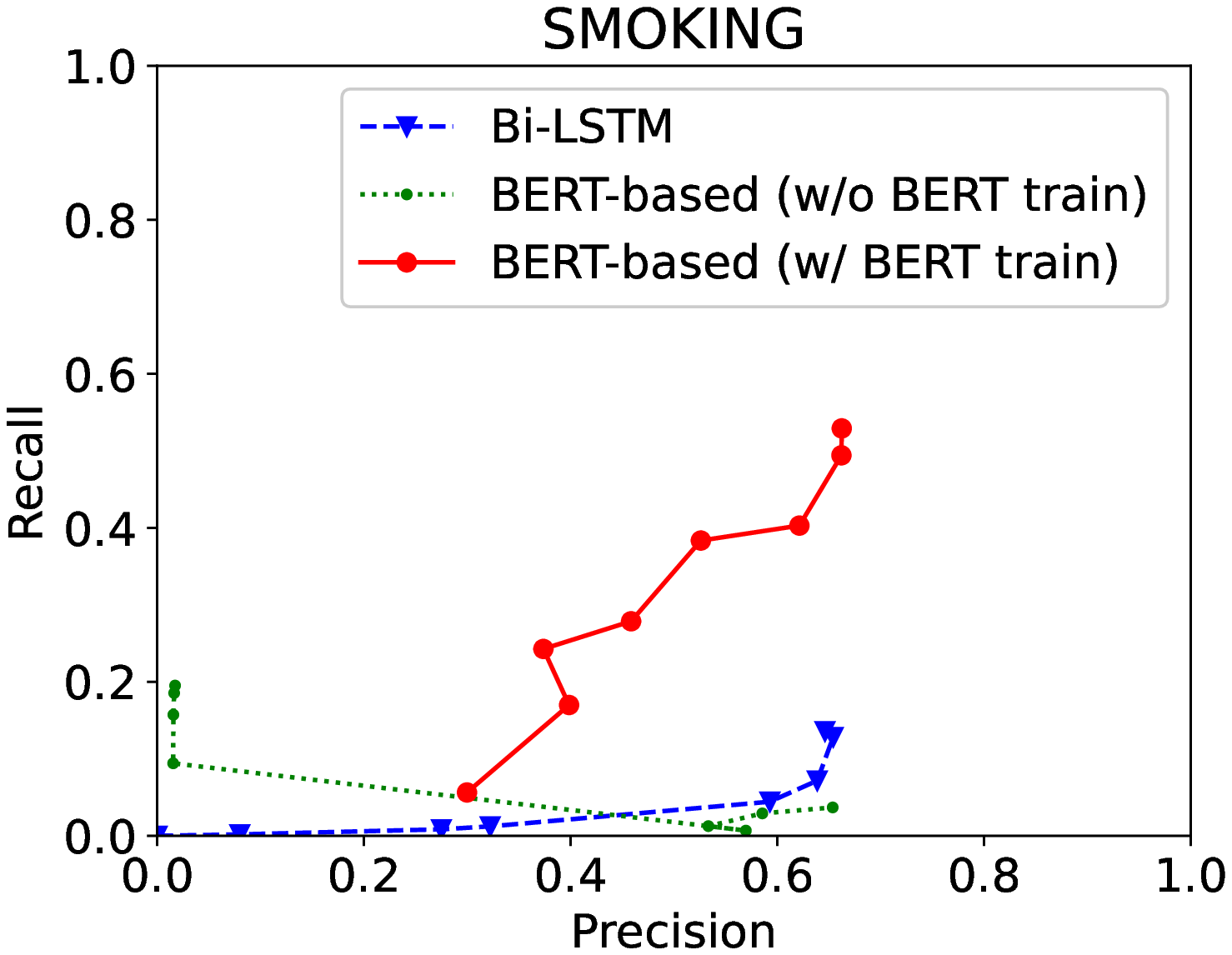}}
\end{picture}
\caption{Precision Recall Curve: in-domain test data (average over five runs).}\label{fig:pr_in_domain}
\end{center}
\end{figure*}

\begin{figure*}[t]
\centering
\begin{center}
\begin{picture}(130, 177)
    \put(-177, 0){\includegraphics[scale=0.55]{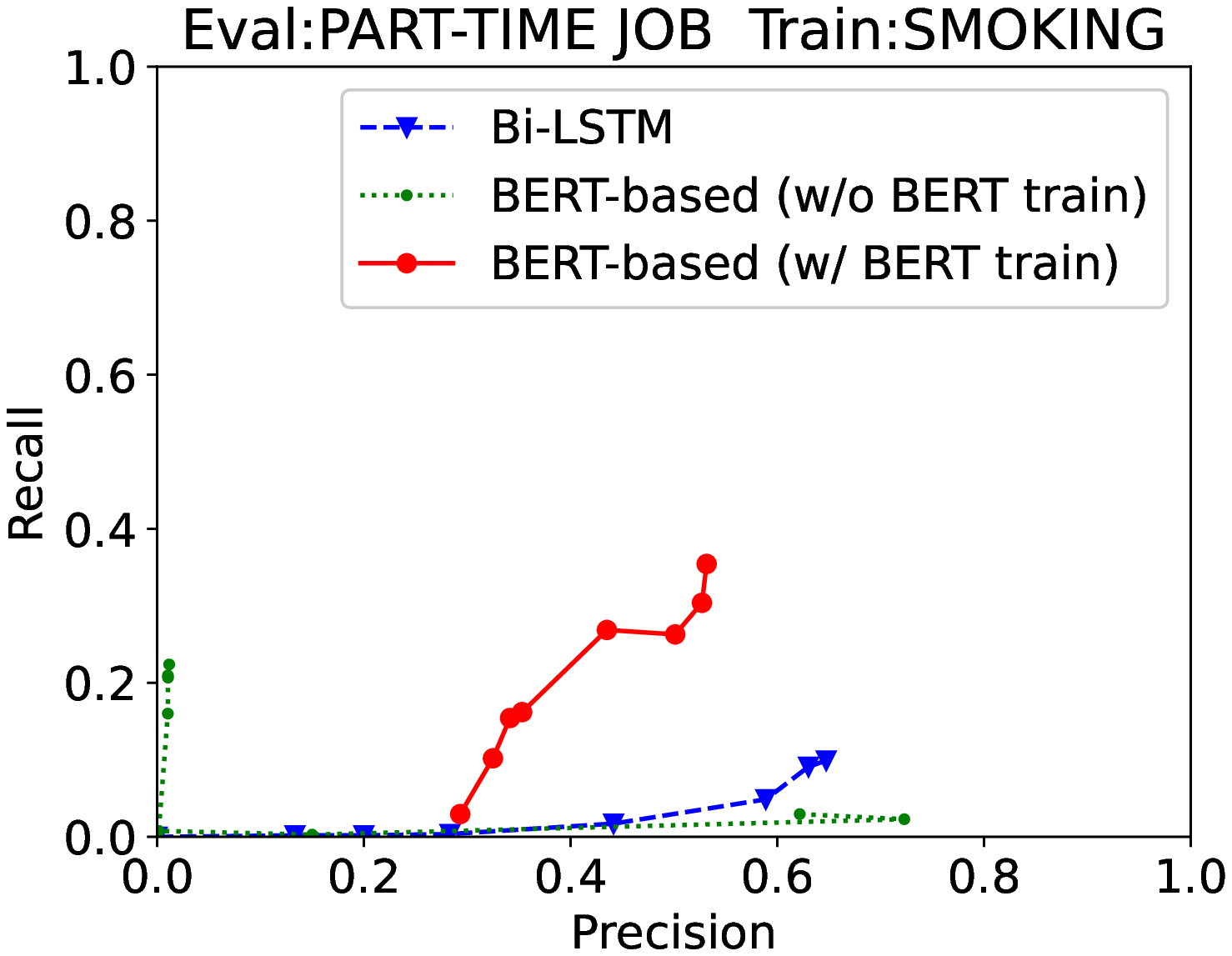}}
    \put(70, 0){\includegraphics[scale=0.55]{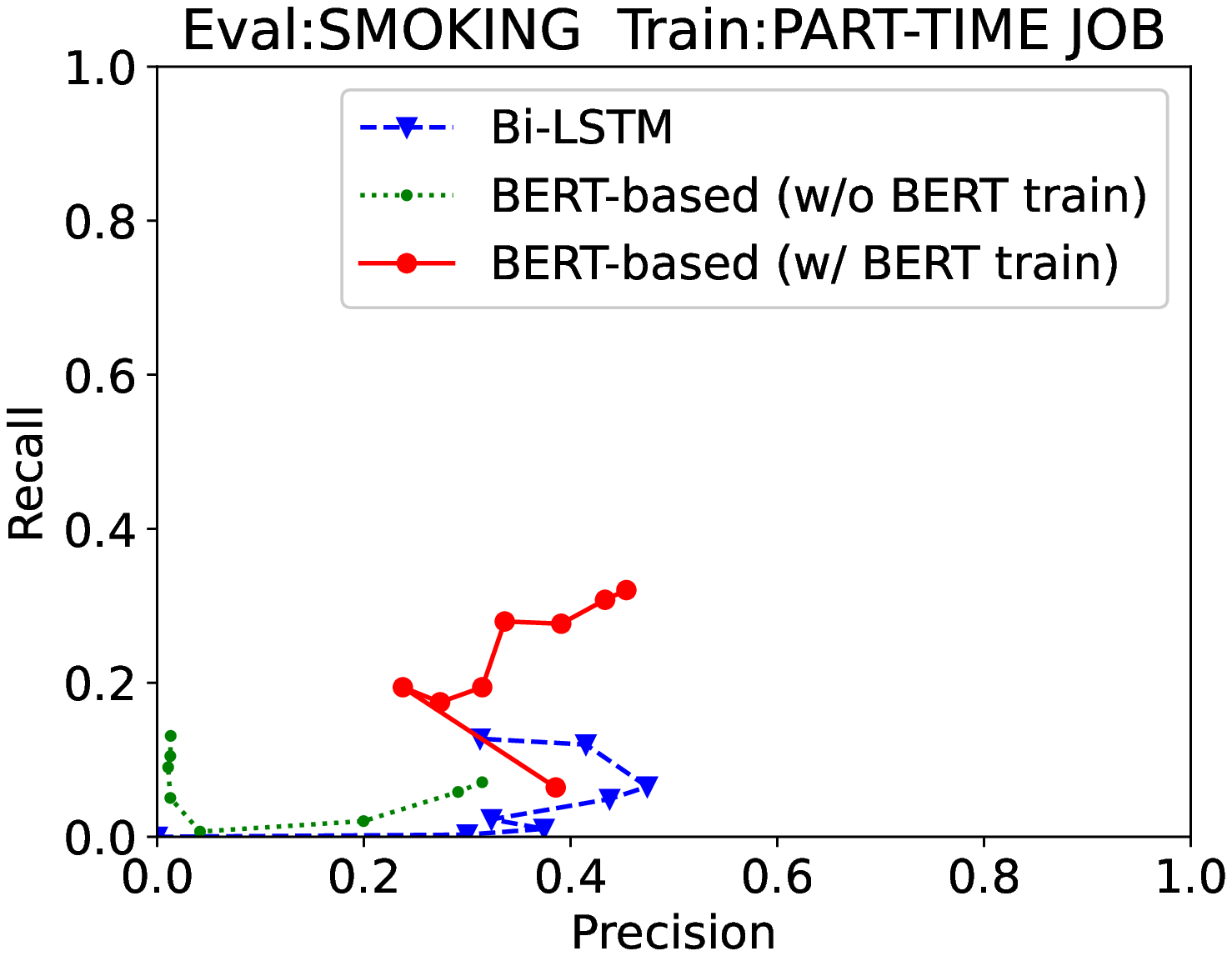}}
\end{picture}
\caption{Precision Recall Curve: out-domain test data (average over five runs).}\label{fig:pr_out_domain}
\end{center}
\end{figure*}

  To look into these points, let us consider precision-recall curves shown in Figure~\ref{fig:pr_in_domain} (in-domain test) and Figure~\ref{fig:pr_out_domain}. Interestingly, all figures show that the BERT-based and BiLSTM-based methods both quickly improves in precision as the number of training sentences available increases while only does the BERT-based method so in recall. In other words, only the BERT-based method learn to recognize various error types with little exposure to error examples. This is surprising because BERT is only pre-trained on a native corpus that are virtually error-free and thus it knows nothing about grammatical errors learners make. Nevertheless, it quickly learns to recognize various error types by fine-tuning with few training instances.

  By contrast, the BERT-based method without BERT train improves either in recall or in precision but not in both. This is probably because it requires much more degree of freedom in terms of the network parameters to learn rules for detecting a wide variety of grammatical errors, which have a certain degree of complexity. Considering the fact that the exact same information about errors (i.e., training data) is given to the three methods, these results lead us to the hypotheses that the BERT-based method uses the knowledge about canonical English (what correct English sentences should look like) and transforms it into rules for detecting grammatical errors by fine-tuning. We will explore these points in detail in the following section.

  These results also shed a light on an important aspect of the BERT-based method in grammatical error detection in practice. Namely, a cost-effective way of developing an error detection system would be to create around 1,000 training sentences for each essay topic; according to Figure~\ref{fig:f1_in_domain}, the gain would be much smaller after 1,000 training sentences. Of course, the results are only for two essay topics and the target errors are limited to errors involving preposition use. Also, no one knows how differently the performance curves grow with a much larger set of training instances. It will be interesting to investigate these points for the future work.%

\section{Looking into Potential of Language Model-based Method with Pseudo Error Data}\label{subsec:effects_on_generation}
  In the previous section, we have seen that the BERT-based method has a much higher generalization ability in grammatical error detection. To look into this phenomenon, we now turn to detection performance of the BERT-based method on the pseudo error data. As describe in Subsect.\,\ref{subsec:data}, we train it on the ten sets of training data and test the trained models on the fixed test set.%

\begin{figure}[t]
\centering
\begin{center}
\begin{picture}(130, 177)
    \put(-57, 0){\includegraphics[scale=0.97]{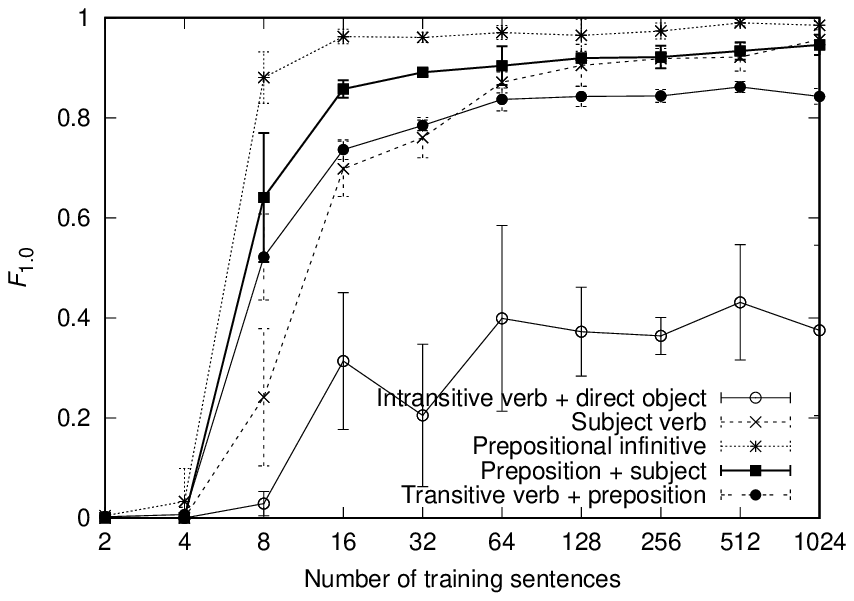}}
\end{picture}
\caption{Detection Performance ($F_{1.0}$) per Error Type in Relation with Training Size: Pseudo Error Data (average over five runs).}\label{fig:nyt_by_nyt_f1}
\end{center}
\end{figure}%

  Figure~\ref{fig:nyt_by_nyt_f1} shows the relationship between the size of training data and $F_{1.0}$ for each error type where the size is measured by the number of sentences. Figure~\ref{fig:nyt_by_nyt_f1} reveals that the BERT-based method already recognizes some of the target errors at early stages of the graph (even with two or four training sentences). Performance goes much higher even with eight training sentences in most of the error types with an exception of the error type ``Intransitive verb + object''. For instance, the BERT-based method recognizes more than half of the ``Preposition + subject'' errors with a precision of 0.800 only with eight training instances. This implies that BERT has certain knowledge of English grammar similar to the notions of POS such as verbs and syntactic relations such as subjects; otherwise, it would be difficult to achieve a similar performance in this type of error considering that the noun phrase of a subject and its position in the sentence considerably vary depending on the target sentence.

  We can make the same argument about the transitive verb + preposition and intransitive verb + object error types. It should be emphasized that the BERT-based method has to detect errors in the verbs that never\footnote{Strictly, some of the verbs may appear in the training sentences for the other error types. However, they never appear in the erroneous phrases. Also, they do not appear at all when the training size is small.} appear in the training data; recall that there is no overlap of the target transitive/intransitive verbs in the training and test data as described in Subsect.\,\ref{subsec:data}. In other words, the BERT-based method can recognize unseen erroneous combinations, for example, \textit{*visited in Atlanta} and \textit{*specialized environmental litigation} after just seeing \textit{*mention in}, \textit{discussed about} (transitive verb + preposition type) and \textit{*were related drugs} and \textit{belongs Lon's grandmother} (intransitive verb + direct object error type). The training and test sentences have almost nothing in common except that they are the combinations of transitive/intransitive verbs and prepositions/objects. Besides, the fact that combinations of other verbs and prepositions/objects that are correctly used often appear in the test data makes the task even more difficult without the knowledge of POS and syntactic relations. These findings support the hypotheses that BERT has grammar-like knowledge and that it can turn the knowledge into error detection rules by fine-tuning.

\section{Exploration for Cost-Effective Error Detection with Feedback Comments}\label{subsec:improvement}
  The findings we have obtained so far bring out the possibility that one can implement with few training instances a system that accurately detects grammatical errors and recognizes their detailed error types. For example, manually or automatically, creating few instances of the erroneous combination of transitive verbs and prepositions as we saw in the previous sections (e.g., \textit{*discuss about}), one can develop a system detecting the same type of error in other transitive verbs and prepositions (e.g., \textit{*mention about it} and \textit{*attend in}). With the detailed error types, the system can also output feedback comments to the user such as in \textit{Transitive verbs do not take a preposition. Instead, they take a direct object} instead of just indicating them as preposition errors.

  As a pilot study, we trained the BERT-based method on the pseudo data and tested it on the real (learner) data to examine the above possibility. To achieve it, we manually annotated the real data with the target five error types consulting the feedback comment attached to each error.

\begin{figure}[t]
\centering
\begin{center}
\begin{picture}(130, 177)
    \put(-57, 0){\includegraphics[scale=0.97]{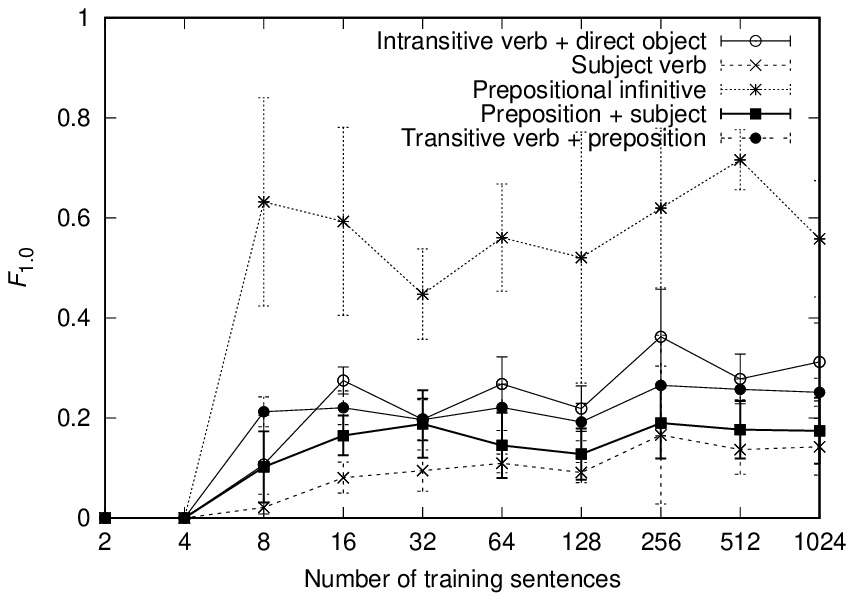}}
\end{picture}
\caption{Detection Performance ($F_{1.0}$) on per Error Type in Relation with Training Size: Trained on Pseudo Error Data; Tested on Real Data (PART-TIME JOB and SMOKING merged) (average over five runs).}\label{fig:both_by_nyt_f1}
\end{center}
\end{figure}

  Figure~\ref{fig:both_by_nyt_f1} shows the results. Figure~\ref{fig:both_by_nyt_f1} reveals that the BERT-based method on the pseudo data does not perform on the real data as well as on the pseudo data. Performance growths stop at an early stage (around eight training sentences).

  A possible reason for this is that in the real data, multiple errors often appear in a sentence. Also, multiple errors in a sentence can range over multiple types of error. Besides, the error rate is much lower in the real data than in the pseudo error where one error occurs per sentence except 200 error-free sentences (although multiple types of error appear in the whole data set). These conditions make the task much more difficult in the real data.%

  Having said that, the results shown in Figure~\ref{fig:nyt_by_nyt_f1} still encourages us to develop language model-based systems with a small amount of in-domain training data in order to detect grammatical errors with detailed error types. One possible way to achieve it is (i) to sample sentences from unannotated essays written on the target topic; (ii) to annotate them with the specific types errors that the developer wants to give feedbacks to the user. This will naturally mitigate the problems caused by the multiple-type multiple error situation and the error rate difference. One can also manually create sample error sentences (and their correct versions) to augment the data created by (i) and (ii).

\section{Conclusions}\label{sec:conclusions}
  In this paper, we have explored the capacity of a large-scale masked language model to recognize grammatical errors. Our findings are summarized in the following three points: (1) Experiments with the real learner data show that a BERT-based error detection method has a much higher generalization ability in grammatical error detection than a non-language model-based method and the first performance saturation comes at the point of around 1,000-2,000 training instances; (2) It starts to recognize the target errors with few (as few as two) instances of them; (3) The high generalization ability brings out its potential for developing systems that detect and explain grammatical errors with very few training instances.


\bibliographystyle{acm}
\bibliography{paper}

\end{document}